\documentclass[10pt,twocolumn,letterpaper]{article}

\usepackage{icb}
\usepackage{times}
\usepackage{epsfig}
\usepackage{graphicx}
\usepackage{amsmath}
\usepackage{amssymb}

\usepackage{verbatim}
\usepackage{booktabs} 
\usepackage{adjustbox}
\usepackage{colortbl} 
\usepackage{xcolor} 
\usepackage{xfrac}
\usepackage[]{algorithm2e}
\usepackage{subcaption}
\usepackage{graphicx,multirow}
\usepackage{array}

\usepackage{epsfig}
\usepackage{graphicx}
\usepackage{verbatim}
\usepackage{booktabs} 
\usepackage{colortbl}
\usepackage{mathtools}
\usepackage[norule,symbol,perpage]{footmisc}

\DeclareMathOperator{\sign}{sign}

\icbfinalcopy % *** Uncomment this line for the final submission

\ificbfinal\fi

\makeatletter 
\def\ps@IEEEtitlepagestyle{ 
\def\@oddfoot{\mycopyrightnotice} 
\def\@evenfoot{} 
} 
\def\mycopyrightnotice{ 
{\hfill \footnotesize 978-1-7281-3640-0/19/\$31.00 \copyright 2019 IEEE\hfill} 
} 
\makeatother 
\begin{document}

%%%%%%%%% TITLE
\title{Adversarial Examples to Fool Iris Recognition Systems}

\author{Sobhan Soleymani, Ali Dabouei, Jeremy Dawson, and Nasser M. Nasrabadi, {\it Fellow, IEEE}\\
West Virginia University\\
%Institution1 address\\
{\tt\small {\{ssoleyma, ad0046\}@mix.wvu.edu,}}
{\tt\small {\{jeremy.dawson, nasser.nasrabadi\}@mail.wvu.edu}}}

\maketitle
\thispagestyle{empty}

%%%%%%%%% ABSTRACT
\begin{abstract}
Adversarial examples have recently proven to be able to fool deep learning methods by adding carefully crafted small perturbation to the input space image. In this paper, we study the possibility of generating adversarial examples for code-based iris recognition systems. Since generating adversarial examples requires back-propagation of the adversarial loss, conventional filter bank-based iris-code generation frameworks cannot be employed in such a setup. Therefore, to compensate for this shortcoming, we propose to train a deep auto-encoder surrogate network to mimic the conventional iris code generation procedure. This trained surrogate network is then deployed to generate the adversarial examples using the iterative gradient sign method algorithm~\cite{kurakin2016adversarial}. We consider non-targeted and targeted attacks through three attack scenarios. Considering these attacks, we study the possibility of fooling an iris recognition system in white-box and black-box frameworks. 
\end{abstract}
%\let\thefootnote\relax\footnotetext{\mycopyrightnotice} 
%%%%%%%%% BODY TEXT
\section{Introduction}
Biometric systems are widely deployed in many recognition and security applications. Iris images provide the most reliable human identification trait, since,  due to the chaotic morphogenesis involved in the formation of the iris pattern, there is a very large variability of iris patterns among different persons~\cite{daugman2009iris}. Additionally, although externally visible, the iris is  thought to be relatively stable over the time since it is well-protected as an internal organ. However, the recognition performance of raw iris images can be reduced in imaging instances that include cases such as light reflections from the eye's surface, occlusions and fluctuations of perspective, and illumination~\cite{szewczyk2012reliable}. 

The majority of iris recognition systems generate the iris template from the eye image. This process  consists of filters such as 2-D Gabor and wavelets. The binarized iris template is considered as the iris-code, which removes the unwanted amplitude information. Furthermore, storing the iris-code in the database is more secure than the actual eye image, since the eye image can provide sensitive information, including health data, about the subjects. In addition, storing only the iris-code is sufficient for future identity verification attempts~\cite{venugopalan2011generate}. However, iris biometric recognition systems are vulnerable to a diverse set of attacks~\cite{raghavendra2014presentation}. Presentation attacks on iris recognition systems are well-studied in the biometrics literature~\cite{raghavendra2015robust}. These  attacks undermine the performance of a recognition system by presenting biometrics that are similar to those of an authorized user. 

\begin{figure}[t]
\begin{center}
\includegraphics[width=1\linewidth]{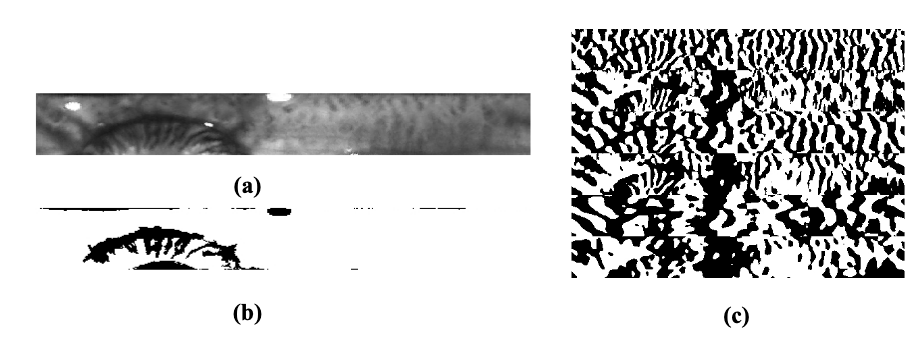}
\end{center}
\caption{(a) Normalized iris image, (b) normalized mask, and (c) the corresponding iris code generated using~\cite{krichen2008osiris}.}
\label{fig:iris}
\end{figure}
{\it Adversarial examples}, introduced in~\cite{kurakin2016adversarial}, are samples of input data modified such that machine learning classifiers are fooled. While, in many cases, these modifications are perceptually indistinguishable and cannot be noticed by a human observer~\cite{kurakin2016adversarial,carlini2017adversarial}. Adversarial examples can be utilized to conceal the identity of a subject or fool the security system to provide access to an unauthorized subject by matching the adversarial example to a specific or any other authorized subject. Adversarial examples are considered  security threats since an adversarial example that is designed to be misclassified by one model is often also misclassified by the other models~\cite{bruna2013intriguing}. Therefore, adversarial examples can be generated without the exact knowledge of the recognition framework. The majority of the attacks applied on iris are done in the iris code domain~\cite{rathgeb2017feasibility}. However, spatial correlation between the adjacent locations in the iris and the nature of the filters used in iris-template generation, result in correlation between iris-code bits~\cite{daugman2016information,hollingsworth2009best}. Therefore, the modification of the bits in the iris-code domain does not necessarily represent a real physical iris image.  

The authors in~\cite{rathgeb2017feasibility} have tried to solve this problem by altering the unstable bits. They defined the unstable bits as the bits between consecutive 1-bit and 0-bit sequences. However, this cannot be an optimal solution, since depending on the filters used in iris template generation, this correlation can occur in adjacent bits, non-adjacent bits, or even non-adjacent rows. For example, when the template generation process consists of a bank of Gabor filters~\cite{krichen2008osiris}, locations far away from each other can be highly correlated. Therefore, altering the bits in the iris-code may result in a code that does not represent an actual iris. On the other hand, although there are attempts to generate the natural iris image from a given iris code using evolutionary algorithms~\cite{galbally2013iris}, this is a very complex task since, theoretically, it is impossible to generate the iris image from the iris code~\cite{czajka2018presentation}.

In this paper, we make the following contributions: i) we mimic the iris-code generation filter bank procedure with a surrogate deep network, ii) the surrogate deep network is differentiable with respect to its parameters and its input, which is then used to generate adversarial examples, iii) the possibility of generating adversarial irises when the attacker is provided only with a partial knowledge of the iris recognition model is investigated, and iv) several attack scenarios are examined for non-targeted and targeted frameworks. The non-targeted frameworks investigate the possibility of fooling the iris recognition system not to recognize the identity of the subject. While, in the targeted framework the attacker attempts to represent a specific subject. 
\section{Related Works}
\subsection{Iris Code}
The iris modality is among the most promising biometric features that illustrates high-performance with a reasonable confidence level. The majority of iris recognition systems utilize the iris template generated from the eye image. However, the performance of iris recognition systems become limited when the amplitude of the iris template is considered in the recognition algorithm, because the amplitude is sensitive to the light reflections, occlusions, illumination change. Therefore, most of the iris recognition frameworks in practice are based on the phase information~\cite{szewczyk2012reliable}, where to bypass the amplitude effect, the iris template is binarized. Binarizing the iris template, which is refereed to as the iris-code, removes the effect of the amplitude information. 

The iris-code is generally constructed through segmentation, mask generation, normalization, and binerization. To generate the iris-code, the eye image is segmented to find the iris image to identify the area of interest. Segmentation is performed by finding the pupillary and the limbus boundaries of the iris. The mask image is then generated using the segmented eye image. In the binary mask, the ones represent the iris. This binary mask is used during the matching step in order to ignore noisy pixels and the pixels that do not belong to the iris. 
The iris and mask images are then normalized into a rectangular shape following rubber sheet model~\cite{daugman2009iris}. The normalized iris image is finally converted to an iris template through multiple levels of 2-D Gabor or wavelets filters. To generate the iris-code, the iris template is binarized. The iris-code is used in combination with the normalized mask image to verify or classify the iris images. In the authentication or recognition phase, iris-codes are compared using bit-based metrics such as Hamming distance. Several prominent iris authentication frameworks are built through this general framework~\cite{daugman2009iris,szewczyk2012reliable,krichen2008osiris,masek2003matlab}. Figure~\ref{fig:iris} presents a normalized iris image, the corresponding normalized mask, and the corresponding iris-code generated using algorithm described in~\cite{krichen2008osiris}. 
\begin{figure*}
\begin{center}
\includegraphics[width=0.95\linewidth]{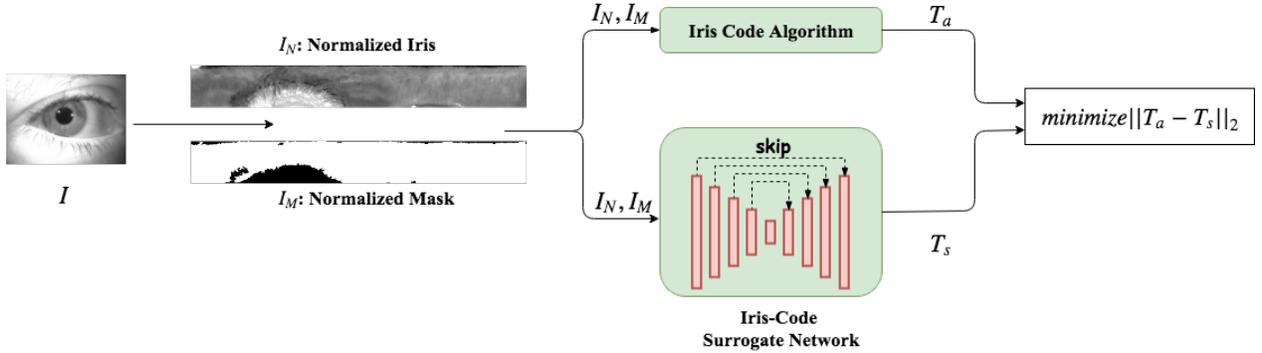}
\end{center}
\caption{Iris code generation: The normalized iris image and the normalized mask are concatenated in depth as the input to the iris code surrogate deep network, while, the output is forced to mimic the iris code by minimizing the reconstruction loss.}
\label{fig:surrogate}
\end{figure*}
\subsection{Iris Presentation Attacks}
Presentation attacks make a fake presentation to the sensor in order to fool the recognition system and cause it to make an incorrect decision~\cite{czajka2018presentation}. In this scenario, the attacker needs the subject's iris image, iris code, or equivalent information. Presentation attack technologies can coarsely be categorized into technologies associated with artifacts and the actual eye. Well-known examples of the first category are paper printouts, textured contact lenses, displays, and prosthetic eyes. The second category includes attacks such as non-conformant use, cadavers, and coercion. 
Paper printouts refer to a printed artifact. In this attack, a hole is cut in the area where the pupil is printed to mimic the reflections typically created capturing an iris image using a camera or illuminator system~\cite{pacut2006aliveness}. Textured contact lenses are manufactured to have a visual texture. The authors of~\cite{baker2010degradation} have shown that, not only the textured contact lenses, but also the clear contact lenses with no visible texture can fool iris recognition systems. Screens can be utilized to display an iris image or video to the sensor as a presentation attack~\cite{he2016multi}. Prosthetic eyes, although requiring a huge amount of time and expertise, have proved to be successful spoofs~\cite{czajka2018presentation}.

Presentation attack detection algorithms are coarsely categorized into hardware-and software-based solutions.   Hardware-based solutions add additional components to the sensor to detect attacks to the sensor. Although these frameworks present better performance, they are more expensive compared to the software-based techniques~\cite{raghavendra2015robust}. Additionally, software-based approaches are more convenient since they are non-invasive, fast, and user-friendly. These techniques identify the attacked image after it is captured by analyzing  its statistical characteristics. On the other hand, the attack detection algorithms can consider the iris as a static or dynamic object~\cite{czajka2018presentation}, while these algorithms can include inducing changes to the iris image. 
\subsection{Adversarial Attacks}
Recently, deep learning models have outperformed the classical machine learning models in a variety of application areas, including biometrics~\cite{soleymani2018multi,soleymani2018generalized,soleymani2018prosodic}. However, these models are vulnerable to a small perturbation in the input image. These small perturbations cannot typically be noticed by a human observer  but can still alter the predictions of the model. Adversarial attacks~\cite{kurakin2016adversarial} attempt to generate adversarial samples, that are very similar to the benign samples, which are misclassified by the classifier~\cite{kurakin2016adversarial,carlini2017adversarial}. When used to attack a biometric security system, these attacks can conceal the identity of a subject or fool the security system to give access to an unauthorized subject.

Adversarial attacks are generally categorized based on the perturbation type they utilize. 
The authors in~\cite{szegedy2013intriguing} introduced a L-BFGS method to generate one of the first adversarial attacks. Although computationally expensive, this method is able to fool deep networks trained on different inputs~\cite{dabouei2019fast}. The Fast Gradient Sign Method (FGSM)~\cite{kurakin2016adversarial} is a fast and  efficient attack based on the sign of the gradient of the classification loss with respect to the input sample as the perturbation. Several extensions to this attack are developed in the literature. The authors in~\cite{rozsa2016adversarial} have proposed to use the gradient value instead of the gradient sign to increase the effectiveness of the attack. Utilizing a Jacobian matrix of the prediction of classes with regards to the input pixels is considered in~\cite{papernot2016limitations} to reduce the number of pixels that are needed to be altered during the attack by calculating the saliency map of the input space. Although this attack requires a very small number of pixels to be modified, saliency-based methods are computationally expensive due to the greedy search for finding the most significant areas in the input sample. In conclusion, these attacks manipulate the classifier by adding high-frequency components to the input sample and using an $L_p$ norm constraint to control the amount of distortion.
\section{Approach}
In this section, we describe the proposed framework, the networks, the architectures, and the attack scenarios. 
\subsection{Problem Statement}
\begin{figure*}
\begin{center}
\includegraphics[width=0.95\linewidth]{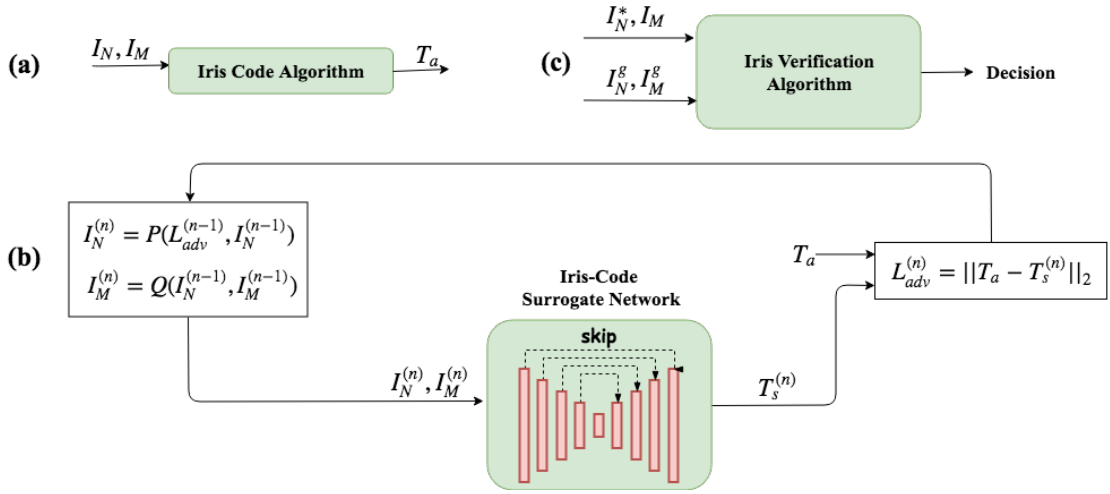}
\end{center}
\caption{Iris adversarial example generation network: (a) The conventional iris-code algorithm is utilized to generate the benign example, (b) The iris-code surrogate network transforms the benign example to generate the adversarial examples by enforcing the adversarial loss. (c) The generated adversarial example is compared to a ground truth example from the gallery.}
\label{fig:1CNN}
\end{figure*}

Spatial correlation between the adjacent locations in the iris and the nature of the filters used in iris template generation, result in correlation between iris-code bits. Depending on the filters that are used in the template generation, this correlation can be between the adjacent or non-adjacent bits. For instance, in the case of the Gabor filter bank used in~\cite{krichen2008osiris}, locations far away from each other can be highly correlated. In addition, generating the natural iris image from the iris-code is computationally expensive. Therefore, directly altering the bits in the iris-code domain cannot be an optimal solution to generating a physically feasible adversarial iris example. 
To address this problem, we mimic the iris-code generation with a deep auto-encoder surrogate network under the reconstruction loss, as shown in Figure~\ref{fig:surrogate}. The trained iris-code surrogate network is then utilized to generate the adversarial examples as shown in Figure~\ref{fig:1CNN}(b). In the paper, OSIRIS iris verification algorithm~\cite{krichen2008osiris} is considered as the conventional iris verification framework. It worths mentioning that the proposed framework can be customized to be applied to the other conventional iris-code based iris recognition algorithms~\cite{daugman2009iris,szewczyk2012reliable,masek2003matlab}.

\subsection{Iris-Code Generation}
Since the generation of an adversarial example framework, requires the back-propagation of the gradient of the adversarial loss with respect to the input image, we use an auto-encoder based on the U-net architecture~\cite{ronneberger2015u} to generate the iris-code. Inspired by~\cite{papernot2016distillation,tramer2016stealing}, the {\it iris-code surrogate network} is defined as a surrogate network which generates iris-codes that are similar to the iris-codes that are generated by the conventional algorithm. As shown in Figure~\ref{fig:surrogate}, the input sample for this network is the normalized iris image, $I_N$, concatenated in depth with the binary normalized iris mask, $I_M$. The output of the network is the generated iris-code, $T_s$. This output is forced to be close to the iris-code generated by the conventional algorithm, $T_a$, using the following reconstruction loss: 
\begin{equation}
\small
L_{rec}=||T_a-T_s||_2.
\end{equation}
The details for the U-Net network used in our experiments can be seen in Table~\ref{table:architecture_table}. The encoding and decoding sub-networks are trained with $2\times 2$ and $1\times 1$ stride sizes, respectively. The output layers of the encoding layers are concatenated in depth with the corresponding layers in the decoding sub-network. Separable kernels~\cite{szegedy2015going} are considered for all the layers. The network is trained using batch size $64$. Batch normalization is applied on the outputs of all the layers. A ReLU activation function is utilized for all the layers except the deconv0 layer, where $tanh$ is considered. Finally, $64{\times} 512{\times} 6$ output is reshaped to $384 \times 512$ to be compatible to the iris-code.

Through the experimental results we can observe that the proposed surrogate network can very closely mimic the conventional iris-code generation algorithms. The error rate for the trained iris-code surrogate network is less than $2\%$, which presents, the generated iris-code, $T_s$, to be sufficiently similar to the conventional iris-code, $T_a$. Since the output of the $tanh$ activation function is in the $[-1,1]$ range, prior to calculating $L_{rec}$, the output is normalized to $[0,1]$, in order to have the same range as $T_a$.
\begin{table}[t]
\caption[Table caption text]{Iris-code deep surrogate network: The first five layers represent the encoding sub-network, while the next five layers are the decoding layers. Conv and deconv represent convolutional and deconvolutional layers, respectively.}  
\begin{center}
\small
\addtolength{\tabcolsep}{-5pt}
\begin{tabular}{l@{\hskip .05in}c@{\hskip .05in}c@{\hskip .05in}c}
\hline
layer&kernel&input&output\\
\hline
conv1&   $4{\times} 4 {\times} 64$ & $64{\times} 512 {\times} 2$& $32{\times} 256{\times} 64$\\
\rowcolor{black!10}conv2&   $4{\times} 4 {\times} 128$& $32{\times} 256 {\times} 64$&$16{\times} 128 {\times} 128$\\
conv3&   $4{\times} 4 {\times} 256$& $16{\times} 128 {\times} 128$& $8{\times} 64 {\times} 256$\\
\rowcolor{black!10}conv4&   $4{\times} 4 {\times} 512$& $8{\times}  64  {\times} 256$& $4{\times} 32 {\times} 256$\\
conv5&   $4{\times} 4 {\times} 512$& $4{\times}  32  {\times} 256$& $2{\times} 16 {\times} 512$\\
\hline
\rowcolor{black!10}deconv4& $4{\times} 4 {\times} 512$& $2{\times}  16  {\times} 512$& $4{\times} 32 {\times} 512$\\
deconv3& $4{\times} 4 {\times} 256$& $4{\times}  32 {\times} (512+256)$& $8{\times} 64 {\times} 256$\\
\rowcolor{black!10}deconv2& $4{\times} 4 {\times} 128$& $8{\times} 64 {\times} (256+256)$&$16{\times} 128 {\times} 128$\\
deconv1& $4{\times} 4 {\times} 64$ & $16{\times} 128 {\times} (128+128)$& $32{\times} 256{\times} 64$\\
\rowcolor{black!10}deconv0& $4{\times} 4 {\times} 6$ & $32{\times} 256 {\times} (64+64)$& $64{\times} 512{\times} 6$\\
\bottomrule
\end{tabular}
\end{center}
\label{table:architecture_table}
\end{table}
\subsection{Adversarial Examples}
After training the iris-code surrogate network to learn the iris-code generation, the trained surrogate network is utilized to generate the adversarial examples. It should be noted that the weights and the parameters of this network are not trained in this adversarial setup. Here, the input to the surrogate network is the normalized iris image, $I_N$, concatenated in depth with the normalized iris mask, $I_M$. In this setup, as can be seen in Figure~\ref{fig:1CNN}(b), at each iteration $n$, both the adversarial normalized iris image, $I_N^{(n)}$, and normalized mask, $I_M^{(n)}$, are updated through enforcing the adversarial loss, $L^{(n)}_{adv}$:
\begin{equation}
\small
L_{adv}^{(n)}=||T_a-T_s^{(n)}||_2,
\label{eq:adv}
\end{equation}
where $T_s^{(n)}$ is the adversarial iris-code generated at the $n^{th}$ iteration. The loss function is  only calculated on the bits which are not already flipped, i.e., locations where the binarized values of $T_s^{(n)}$ are equal to the values of $T_a$. The adversarial setup to update $I_N^{(n)}$ follows a clipped iGSM algorithm~\cite{kurakin2016adversarial}:
\begin{equation}
\small
P: I_N^{(0)}=I_N, I_N^{(n)}=Clip\{I_N^{(n-1)}-\epsilon\sign\nabla_{I_N^{n-1}} L_{adv}^{(n)}\}.
\end{equation}
The {\it clip} function thresholds the values to make sure that the values are inside $[0,1]$ range, {\it sign} represents the sign function, and $\epsilon$ indicates step size in each iteration of the adversarial attack. The variable $\epsilon$ limits each pixel's maximum distance between the $n^{th}$ iteration and the $(n-1)^{th}$ iteration adversarial examples. Since the conventional iris-code is a binary image, the normalized mask is also updated during each step. Through these updates, the bits that are already saturated, are added to the normalized mask. 
%\begin{equation}
%Q: I_M^{(0)}=I_M, I_M^{(n)}=I_M^{(n-1)}\cup \{I_N^{(n-1)} \neq I_N\}.
%\end{equation}
\begin{equation}
\small
    Q: I_M^{(0)}=I_M, I_M^{(n)}(i) =
    \begin{cases*}
      0 & if $I_M^{(n-1)}(i)=0$, \\
      0 & if $I_N^{(n-1)}(i)=0 \& I_N(i)\neq 0$, \\
      0 & if $I_N^{(n-1)}(i)=1\& I_N(i)\neq 1$, \\
      1        & otherwise,
    \end{cases*}
\end{equation}
where, $i$ represent the locations in the normalized iris and mask images. The combination of this transformation and computing the adversarial loss function over the bits that are not flipped yet, results in a more accurate and realistic adversarial attack. Adversarial training is continued until the termination criteria is satisfied or the maximum number of iterations is reached. Following the framework in~\cite{rathgeb2017feasibility}, the termination criteria is $HD(I_N, I_N{(n)})>\delta$, where $HD$ represents the Hamming distance and $\delta$ is the recognition threshold. Since a Hamming distance of $0.32$ results in False Match Rate of about $0.0001\%$, we select $\delta=0.32$. 

Finally, the final generated adversarial iris, $I^{*}_N$, along with the normalized iris mask, $I_M$, are compared with a the ground truth iris image, which consists of a normalized iris image, $I_N^g$, and its corresponding normalized iris mask, $I_M^g$, from the gallery. This comparison can be performed using the whole iris-code , which we refer to as the first scenario, or a series of bits from locations in the iris-code. We investigate the scenario in which these locations are known to the attacker in the second scenario. The third scenario examines the performance of the proposed framework when these locations are not revealed to the attacker.
\subsection{Attack Scenarios}
To illustrate the effectiveness of the proposed non-targeted adversarial framework, three attack scenarios are considered. The first attack scenario focuses on the whole adversarial iris-code. In this scenario, Hamming distance between the generated adversarial iris-code, $T_s^{*}$, which corresponds to $I_N^*$ and $I_M$, and the ground truth iris-code, $T_a^g$, which corresponds to $I_N^g$ and $I_M^g$, is considered as the verification criteria. In the second scenario, bits from a set of locations, $v$, known by the attacker in the generated adversarial iris-code, are compared to the bits from the same locations in the ground truth iris-code, using Hamming distance. To train the adversarial example in the second scenario more specifically, the adversarial loss in~(\ref{eq:adv}) is altered to present the distance between bits in these locations in $T_a$ and $T_s^{(n)}$:
\begin{equation}
\small
L_{adv}^{(n)}=||T_a(v)-T_s^{(n)}(v)||_2.
\label{eq:adv_v}
\end{equation}
The third scenario investigates the possibility of generating adversarial examples when the locations of the bits selected to verify the adversarial example are not known by the attacker. In this scenario, the adversarial loss is the same as~(\ref{eq:adv}).
We investigate the possibility of targeted attacks to iris-codes using the same three scenarios. In this framework, the adversarial loss function is altered to force its corresponding iris-code to move closer to the target iris-code,which represents a different subject: 
\begin{equation}
\small
L_{adv}^{(n)}=-||T_a^{tar}-T_s^{(n)}||_2,
\label{eq:adv_tar}
\end{equation}
where $T_a^{tar}$ represents the target iris-code, obtained from the conventional iris-code algorithm. 
\section{Experimental Setup}
In the experimental setup, to evaluate the performance of the proposed algorithm, OSIRIS iris verification algorithm~\cite{krichen2008osiris} is considered as the conventional framework. This algorithm considers a filter bank of six Gabor filters. The normalized iris and mask images are of size $64\times 512$. This algorithm generates binary images of size $384 \times 512$ as the iris-codes. In our framework, ADAM solver for stochastic optimization~\cite{kingma2014adam} is used to train the surrogate network. All the optimizations are conducted using mini-batch of size $64$ and learning rate of $10^{-4}$.
%\subsection{Datasets}
Two dataset are considered in this experimental setup. The iris-code surrogate network is trained on the BioCop dataset~\cite{BIIC}, then the adversarial framework is tested on the BIOMDATA dataset~\cite{crihalmeanu2007protocol}. The iris-code surrogate network is trained using $10,000$ pairs of normalized iris and mask images. The adversarial framework is tested using $3,040$ iris images from $231$ subjects. 
%\subsection{Network Optimization}
\subsection{Results and Discussion }
Table~\ref{table:epsvsit_targeted}(a) presents the average distance and the number of iterations required to generate adversarial examples for the non-targeted framework for three scenarios. The distance is defined as:
\begin{equation}
\small
dist=\frac{||I_N-I_N^{*}||_2}{\sum I_M},
\label{eq:dist}
\end{equation}
where the denominator represents the number of $1$ bits in $I_M$. For the second and third scenarios, $1024$ bits from randomly selected locations are considered. As can be seen in this table, for each scenario, the number of iterations required to generate the adversarial example increases when $\epsilon$ decreases. On the other hand, the distance between these two images decreases when the step size decreases. This inference is due to the fact that smaller step sizes increase the possibility that the perturbations added to a less important pixel can cancel out during multiple steps. 
\begin{table}[t]
\caption[Table caption text]{The average distance between the benign and adversarial examples and the average number of iterations required for the successful non-targeted and targeted attacks.}
\small 
\begin{subtable}[h]{0.5\textwidth}
 
\begin{center}
\caption[Table caption text]{Non-targeted framework}

\addtolength{\tabcolsep}{-5pt}

\begin{tabular}{l|c|c|c|c|c|c}%{l@{\hskip .05in}c@{\hskip .05in}c@{\hskip .05in}c}

 & \multicolumn{2}{c|}{scenario1} & \multicolumn{2}{c|}{scenario2} & \multicolumn{2}{c}{scenario3} \\ \hline
%&scenario $\#1$&&scenario $\#1$&&\\
$\epsilon$&$dist$&$\#itr$&$dist$&$\#itr$&$dist$&$\#itr$\\
\hline
\rowcolor{black!10} 0.03 &	0.003139&	2.08&	0.002689&	2.14&	0.004834&	2.54\\
0.02 &   0.001835&	2.40&	0.001503&	2.70&	0.003255&	3.36\\
\rowcolor{black!10}0.01	&0.000963&	3.76&	0.000627&	4.12&	0.001894&	5.62\\
0.007	&0.000760&	4.90&	0.000450&	5.36&	0.001554&	7.50\\
\rowcolor{black!10}0.005	&0.000640&	6.40&	0.000352&	6.96&	0.001358&	10.06\\
0.002	&0.000480&	14.12&	0.000232&	14.78&	0.001087&	23.32\\
\rowcolor{black!10}0.001	&0.000436&	27.16&	0.000200&	27.44&	0.001015&	45.74\\
0.0007	&0.000426&	38.46&	0.000191&	38.22&	0.000988&	64.68\\
\rowcolor{black!10}0.0005	&0.000419&	53.52&	0.000187&	52.92&	0.000975&	90.24\\
0.0002	&0.000407&	132.34&	0.000180&	130.08&	0.000950&	223.46\\
\rowcolor{black!10}0.0001	&0.000402&	263.78&	0.000177&	258.26&	0.000943&	446.06\\
\end{tabular}
\end{center}
\end{subtable}
\begin{subtable}[h]{0.5\textwidth}

\begin{center}
\caption[Table caption text]{Targeted framework}
\addtolength{\tabcolsep}{-5pt}
\begin{tabular}{l|c|c|c|c|c|c}%{l@{\hskip .05in}c@{\hskip .05in}c@{\hskip .05in}c}

 & \multicolumn{2}{c|}{scenario1} & \multicolumn{2}{c|}{scenario2} & \multicolumn{2}{c}{scenario3} \\ \hline
%&scenario $\#1$&&scenario $\#1$&&\\
$\epsilon$&$dist$&$\#itr$&$dist$&$\#itr$&$dist$&$\#itr$\\
\hline
\rowcolor{black!10}0.03	&0.005166&	2.72&	0.004707&	6.14&	0.008151&	6.80\\
0.02	&0.003557&	3.64&	0.003326&	7.38&	0.005401&	8.22\\
\rowcolor{black!10}0.01	&0.002113&	6.10&	0.001775&	11.38&	0.003273&	12.40\\
0.007	&0.001778&	8.24&	0.001436&	14.70&	0.002781&	16.10\\
\rowcolor{black!10}0.005	&0.001540&	10.96&	0.001244&	18.92&	0.002399&	20.10\\
0.002	&0.001258&	25.56&	0.001007&	40.74&	0.002036&	43.24\\
\rowcolor{black!10}0.001	&0.001166&	49.76&	0.000938&	76.56&	0.001921&	79.72\\
0.0007	&0.001141&	70.54&	0.000937&	109.10&	0.001891&	110.80\\
\rowcolor{black!10}0.0005	&0.001123&	98.24&	0.000925&	150.90&	0.001884&	152.78\\
0.0002	&0.001099&	243.78&	0.000913&	373.41&	0.001875&	372.32\\
\rowcolor{black!10}0.0001	&0.001091&	486.28&	0.000893&	738.46&	0.001885&	740.08\\
\end{tabular}
\end{center}
\end{subtable}
\label{table:epsvsit_targeted}
\end{table}
However, the distance remains almost unchanged for step sizes smaller than $0.001$, while the average number of iterations required to generate the adversarial example increases proportionally with $\epsilon$. In addition, as expected, the average distance in the first scenario is more than the average distance in the second scenario, since the attacker only needs to alter a subset of the bits. On the other hand, the average distance in the third scenario is more than the first two scenarios, since the attacker does not have any knowledge about the locations of the bits considered in the verification algorithm.

\begin{table}[t]
\caption[Table caption text]{The success rate for the first scenario on (a) non-targeted and (b) targeted frameworks for different step sizes when maximum number of possible iterations changes.} 
\small 
\begin{subtable}[h]{0.5\textwidth}
\begin{center}
\caption[Table caption text]{Non-targeted framework}
\addtolength{\tabcolsep}{-5pt}

\begin{tabular}{l|c|c|c|c|c|c|c|c}%{l@{\hskip .05in}c@{\hskip .05in}c@{\hskip .05in}c}

$\epsilon,\#it$&$10$&$20$&$30$&$40$&$50$&$100$&$200$&$300$\\
\hline
\rowcolor{black!10}0.03 &	100&	100&	100&	100&	100&	100&100&100\\
0.02 &  99.4&	99.8&	100&	100&	100&	100&100&100\\
\rowcolor{black!10}0.01	&91.2&	95.7&	100&	100&	100&	100&100&100\\
0.007	&75.4&	93.6&	100&	100&	100&	100&100&100\\
\rowcolor{black!10}0.005	&50.3&	91.3&	97.3&	100&	100&	100&100&100\\
0.002	&2.3&	39.1&	69.4&	89.5&	95.2&	100&100&100\\
\rowcolor{black!10}0.001	&0&	3.5&	15.7&	42.9&	57.3&	98.1&100&100\\
0.0007	&0&	0&	6.1&	13.8&	30.5&	87.1&100&100\\
\rowcolor{black!10}0.0005	&0&	0&	0&	4.5&	8.7&	57.8&96.1&99.6\\
0.0002	&0&	0&	0&	0&	0&	5.3&45.6&74.1\\
\rowcolor{black!10}0.0001	&0&	0&	0&	0&	0&	0&5.2&12.3\\

\end{tabular}
\end{center}
\label{table:epsvsit_2}
\end{subtable}

\begin{subtable}[h]{0.5\textwidth}

\begin{center}
\caption[Table caption text]{Targeted framework}

\addtolength{\tabcolsep}{-5pt}

\begin{tabular}{l|c|c|c|c|c|c|c|c}%{l@{\hskip .05in}c@{\hskip .05in}c@{\hskip .05in}c}

$\epsilon,\#it$&$10$&$20$&$30$&$40$&$50$&$100$&$200$&$300$\\
\hline
\rowcolor{black!10}0.03 &	100&	100&	100&	100&	100&	100&100&100\\
0.02 &  95.1&	97.6&	100&	100&	100&	100&100&100\\
\rowcolor{black!10}0.01	&88.4&	93.1&	100&	100&	100&	100&100&100\\
0.007	&71.1&	88.7&	100&	100&	100&	100&100&100\\
\rowcolor{black!10}0.005	&36.2&	82.7&	94.8&	100&	100&	100&100&100\\
0.002	&1.8&	29.7&	61.5&	89.5&	95.2&	100&100&100\\
\rowcolor{black!10}0.001	&0&	2.1&	11.2&	32.4&	49.5&	95.3&100&100\\
0.0007	&0&	0&	4.3&	10.1&	22.7&	83.7&92.1&100\\
\rowcolor{black!10}0.0005	&0&	0&	0&	2.1&	5.2&	41.2&89.8&97.6\\
0.0002	&0&	0&	0&	0&	0&	4.4&38.7&61.3\\
\rowcolor{black!10}0.0001	&0&	0&	0&	0&	0&	0&3.2&6.5\\

\end{tabular}
\end{center}
\end{subtable}
\label{table:epsvsit_targeted_2}

\end{table}

Table~\ref{table:epsvsit_targeted}(b) presents the results for the targeted framework. To consider a more realistic framework, for each attack, we select the iris images that belongs to the same eye, left or right, to increase the overlap between the normalized iris masks. In this framework the perturbation is added to a benign iris image is order to force it to generate an iris-code close to the conventional iris-code from another subject. Here, we consider $\delta=0.25$ as the termination constraint, while $\delta=0.32$ is considered as the verification threshold. In this framework the distance is defined as: 
\begin{equation}
\small
dist=\frac{||I_N-I_N^{*}||_2}{\sum I_M\cap I_M^{tar}},
\label{eq:dist_tar}
\end{equation}
where $I_M^{tar}$ is the normalized mask image for the target iris image. Here, the denominator represents the number of pixels in which both normalized iris mask images are equal to $1$. As shown in this table, although the number of iterations and the total amount of perturbation required to generate the adversarial example increases compared to the non-targeted framework, the overall trend of the parameters with respect to the step size is similar to the non-targeted framework.  
\begin{figure*}
\begin{center}
\includegraphics[width=0.95\linewidth]{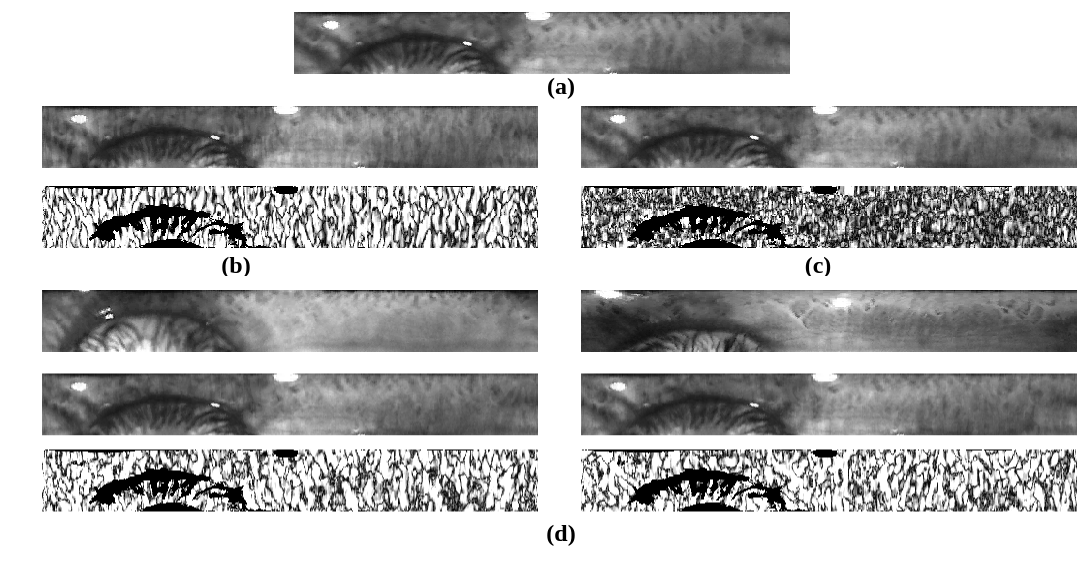}
\end{center}
\caption{(a) The benign example. Adversarial examples and normalized to $[0,1]$ absolute values of perturbation for non-targeted (b) first and (c) second scenarios. (d) Two adversarial examples generated by the targeted first scenario. The benign iris is transformed to the images in the second row and verified as subjects represented by the iris images in the top row.}
\label{fig:results}
\end{figure*}
Table~\ref{table:epsvsit_targeted_2}(a) presents the adversarial attack success-rate for different $\epsilon$ values and the maximum number of iterations in the first scenario for the non-targeted framework. The success-rate is defined as the percentage of successful attacks, given the maximum number of allowed iteration and the randomly chosen ground truth iris image. The maximum number of allowed iterations is directly proportional to the maximum possible difference between the benign and adversarial examples in each pixel. As expected, for small values of the allowed iterations, the success-rate drops drastically when the step size decreases. Table~\ref{table:epsvsit_targeted_2}(b) presents the success-rate for the first scenario in the targeted framework. Compared to the non-targeted framework, the success-rate drops drastically. However, the overall trend of the success-rate is similar to the non-targeted framework. 

Figure~\ref{fig:results} presents benign and generated adversarial examples for targeted and non-targeted frameworks when $\epsilon=0.0002$. As shown in this figure, the adversarial examples are perceptually close to the benign examples, while recognized as a different subject. Figure~\ref{fig:results}(b) and (c) present the adversarial example and normalized to $[0,1]$ absolute value of perturbations for the non-targeted framework using first and second scenarios, respectively. For the second scenario, also the total amount of the perturbation added to the image is less, the perturbation is clustered in certain locations, while in the first scenario, the perturbation is more smoothly distributed.  Figure~\ref{fig:results}(d) presents two examples of the targeted framework using the first scenario. Here, the top iris images are the target images. The second row images, also perceptually very close to the benign image, are verified as the subjects represented by target iris image.

\section{Conclusions}
In this paper, we generated physical adversarial examples for code-based iris recognition systems. However, conventional iris-code generation algorithms are not differentiable with respect to the input image. Generating adversarial examples requires back-propagation of the adversarial loss. Therefore, we proposed to deploy a deep surrogate auto-encoder network to generate iris-codes very similar to iris-codes generated by conventional algorithm. The adversarial network uses the trained surrogate network to generate the adversarial examples using fast gradient sign descent algorithm. We examined the possibility of generating non-targeted and targeted adversarial examples. Considering three white-box and black-box attack scenarios, the proposed network was able to deceive the iris recognition system, while the perturbation added to the benign examples remain in the acceptable range. %In these cases, we examine the possibility of fooling the recognition system whether or not the attacker is provided with the full knowledge of the recognition system. 
\begin{center}
ACKNOWLEDGEMENT
\end{center}
This work is based upon a work supported by the Center for Identification Technology Research and the National Science Foundation under Grant $\#1650474$.
{\small
\bibliographystyle{ieee}
\bibliography{bib}
}

\end{document}